\documentclass[manuscript,screen]{acmart}
\usepackage{xargs}  
\usepackage{caption}
\usepackage{subcaption}
\usepackage{algorithmic}
\usepackage{graphicx}
\usepackage{textcomp}
\usepackage{xcolor}
\usepackage{soul}
\usepackage{gensymb}
\usepackage{siunitx}
\usepackage{textcomp}
\usepackage{multirow} 
\usepackage{makecell}
\usepackage{comment}
\usepackage{pifont}
\usepackage{paralist}
\usepackage{threeparttable}
\usepackage{rotating}
\usepackage{todonotes}
\newcounter{todocounter}

\AtBeginDocument{%
  \providecommand\BibTeX{{%
    \normalfont B\kern-0.5em{\scshape i\kern-0.25em b}\kern-0.8em\TeX}}}

\setcopyright{acmcopyright}
\copyrightyear{2022}
\acmYear{2022}
\acmDOI{XXXXXXX.XXXXXXX}

\begin{document}

\title{\textit{Covy}: An AI-powered Robot with a Compound Vision System for Detecting Breaches in Social Distancing}


\author{Serge Saaybi}
\affiliation{%
  \institution{The Delft University of Technology}
  \city{Delft}
  \country{The Netherlands}}

\author{Amjad Yousef Majid}
\affiliation{%
  \institution{The Delft University of Technology}
  \streetaddress{Mekelweg 4}
  \city{Delft}
  \country{The Netherlands}}
\email{a.y.majid@tudelft.nl}

\author{R Venkatesha Prasad}
\affiliation{%
  \institution{The Delft University of Technology}
  \city{Delft}
  \country{The Netherlands}}
  
\author{Anis Koubaa}
\affiliation{%
  \institution{Prince Sultan University}
  \city{Riyadh}
  \country{Saudi Arabia}}

\author{Chris Verhoeven}
\affiliation{%
  \institution{The Delft University of Technology}
  \city{Delft}
  \country{The Netherlands}}

\renewcommand{\shortauthors}{Amjad Yousef Majid, et al.}

\begin{abstract}
This paper introduces a compound vision system that enables robots to localize people up to $\SI{15}{\meter}$ away using a cheap camera. And, it proposes a robust navigation stack that combines Deep Reinforcement Learning (DRL) and a probabilistic localization method. 
To test the efficacy of these systems, we prototyped a low-cost mobile robot that we call \texttt{Covy}. Covy can be used for applications such as promoting social distancing during pandemics or estimating the density of a crowd.
We evaluated Covy's performance through extensive sets of experiments both in simulated and realistic environments. Our results show that Covy's compound vision algorithm doubles the range of the used depth camera, and its hybrid navigation stack is more robust than a pure DRL-based one.

\end{abstract}

\ccsdesc[500]{Human-centered computing~Empirical studies in ubiquitous and mobile computing}

\keywords{Human-robot interaction, mobile robots, vision, deep reinforcement learning, covid-19}

\maketitle

\section{Introduction}
\begin{figure}
    \centering
    \includegraphics[width=.8\columnwidth]{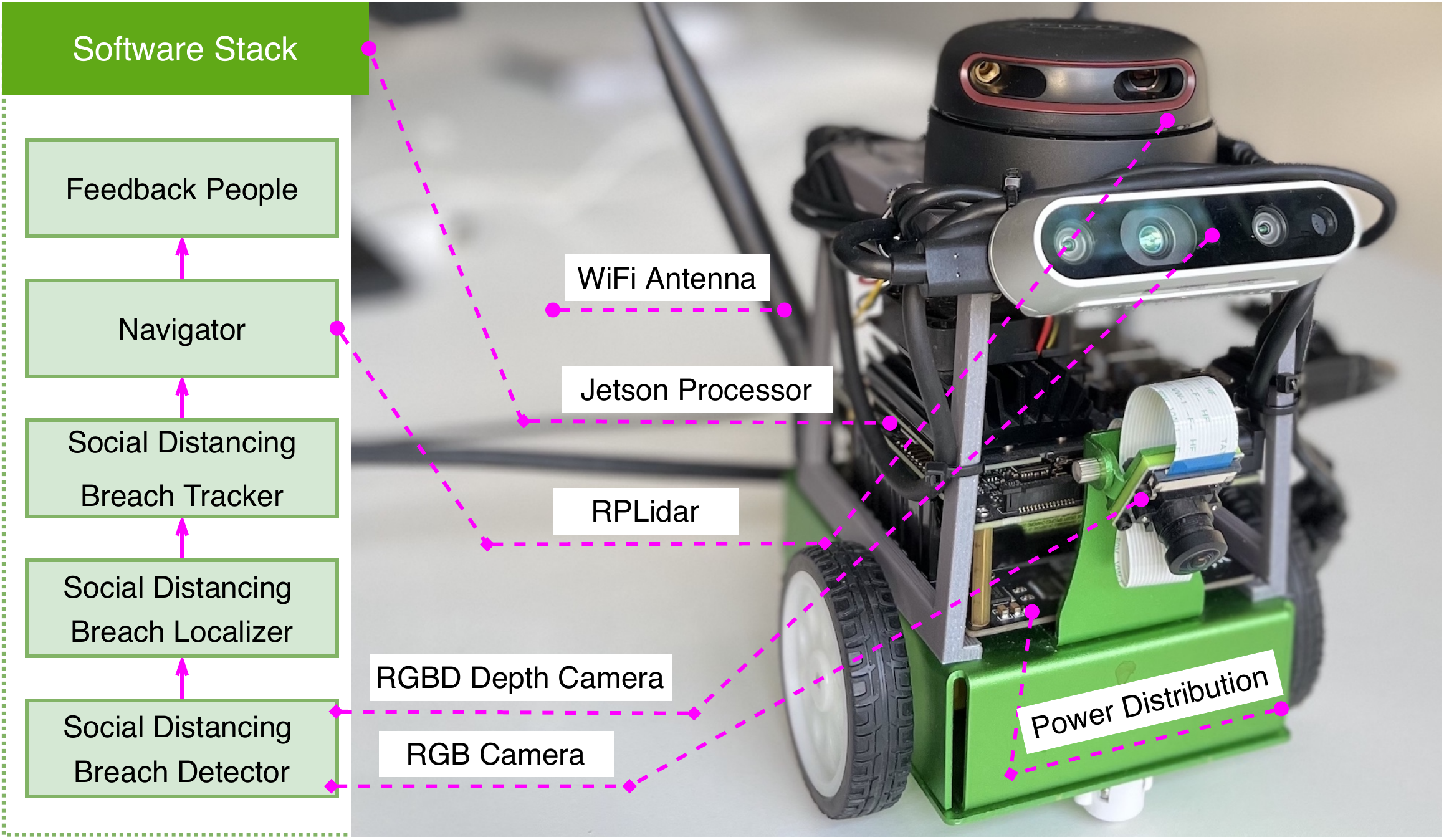}
    \caption{Hardware and software overview of Covy: a prototype swarm robot for promoting social distancing practice.}
    \label{fig:robot_overview}
\end{figure}

Robots are becoming ubiquitous in our daily life (e.g. serving in hotels and restaurants ~\cite{choi2020service}, inspecting plants~\cite{lopez2020mhyro}, etc.). Events like the COVID-19 pandemic have accelerated their adoption in our society. For example, robots have been proposed to encourage people to maintain safe distances and wear masks during the said pandemic~\cite{DeepSOCIAL2020}. Such applications pose an interesting trade-off: \textit{mass production demands the robots to be cheap, but operating in complex environments requires them to be sophisticated, and consequently expensive}. 
Obviously, addressing this challenge can be done in two ways: (i) using an advanced and expensive robot (e.g., Spot~\cite{huang2020agile}) and then trying to bring the cost down for production en masse; or (ii) starting with a cheap robot and then advancing it to meet the application requirements. 
We opted for the latter and developed \textit{Covy} (Figure~\ref{fig:robot_overview}).  Covy is equipped with the Intel RealSense D435i depth camera~\cite{realsense} which has a nominal range of \SI{10}{\meter}. However, its effective range according to our experiments is about \SI{6}{\meter} which is too limited to detect breaches in social distancing---the application that we focused on to test Covy's vision and navigation systems. To navigate to the breachers, Covy relies on its LiDAR and a hybrid navigation stack that combines the Adaptive Monte Carlo Localization (AMCL) algorithm~\cite{amcl} with a DRL agent. Developing a robot like Covy poses many challenges; however, here, we only mention the ones relevant to our research focus.

\noindent\textbf{Challenges.} 
Covy was developed with two main challenges in mind:  

\begin{inparaitem}
\noindent \item[\ding{202}]  Designing a long-range, low-cost vision system that can estimate the 3D coordinates of people in the scene. The localization of nearby people must be sufficiently accuracy to enable safe human-robot interaction. 
\item[\ding{203}] AI algorithms show an excellent ability to generalize across different tasks and hardware. As such, Covy's navigation stack should be AI-based to facilitate code portability and generalizability. 




\end{inparaitem}

\noindent\textbf{Contributions.} Our contributions are manifold.  
\begin{inparaitem}
\item[\ding{202}] We built a platform with complete stack called \textit{Covy} and used it for evaluation.
\item[\ding{203}] We developed a compound vision system that doubles the effective range of the Intel RealSense depth camera for detecting social distancing breaches. 
\item[\ding{204}] We developed a hybrid navigation stack that combines the power of Deep Reinforcement Learning and a probabilistic localization method. 
\item[\ding{205}] We evaluated Covy by conducting extensive experiments in simulated and realistic environments. 
\end{inparaitem}

\section{Related Work}
\subsection{Technology-Driven Social Behavior Encouragement.}
Encouraging desired social behaviors using robots is an emerging field of research that lies on the intersection between swarm robotics~\cite{dorigo2021swarm,bredeche2022social}, Artificial Intelligence (AI)~\cite{russell2010artificial}, and human-robot interaction~\cite{alhafnawi2022mosaix}. 
This emergence has been accelerated during the COVID-19 pandemic as reducing human-to-human interaction was found to be necessary to limit the spread of the virus~\cite{SUN2020102390,Voko2020}. Consequently, many technological solutions have been proposed to keep our society functioning in such situations. 
These proposals can be generally categorized into two categories: passive and active systems. 

Passive systems focus on detecting and analyzing social behaviors~\cite{Rezaei2020,Ahmed2020,punn2021monitoring}. In general, these systems utilize CCTV security cameras and an object detection algorithm (e.g., YOLO~\cite{redmon2018yolov3}) to detect people in the scene. They also leverage tracking algorithms such as SORT \cite{Bewley2016} or DeepSORT \cite{wojke2017simple} to keep track of the identified pedestrians across multiple frames. The obvious limitation of these systems is that they cannot provide feedback to people which is necessary to limit the spread of a virus, for example. 

Active systems, on the other hand, provide feedback to users when they violate the recommended guidelines. They can be further sub-divided into end-user dependent and end-user independent systems. 
Smartphones and wearables are examples of end-user dependent active systems. 
They have been proposed to be used for maintaining safe distances from other people during  COVID-19~\cite{channa2021rise,Sun2020,shen2020robots}. However, privacy concerns discourage people from installing the needed applications which renders this option ineffective. In contrast to smartphones and wearables, robots are end-user independent systems. 
They can be deployed on demand to targeted locations to encourage people to follow certain recommendations. 
Different types of robots have been proposed to support us during the fight against the COVID-19 virus. 
Dr. spot is a teleoperated quadruped robot that monitors social distancing and face masking in public space and alerts offenders ~\cite{shen2020robots}. \citet{fan2020autonomous} introduced a fully autonomous expensive surveillance quadruped for desired social behaviors encouragement. The high prices of these robots make their large-scale deployment prohibitively expensive.
Therefore,  \citet{sathyamoorthy2020covidrobot} proposed a low-cost robotic system -- similar to Covy -- able to detect and navigate towards the breaches using an RGB-D camera and 2D LiDAR. Their system also leverages static CCTV camera to increase the detection range. However, its major drawback is the limited detection range of only up to 4\,m for the mounted depth camera and 3\,m for the fixed CCTV camera. We, in this work, propose a compound low-cost vision system with a breach detection range of up to 16\,m.

\subsection{Robot Navigation.}
There is a growing interest in DRL-based robotic navigation  \cite{tai2017virtualtoreal, surmann2020deep, kahn2018self, majid2021deep}. This is driven mainly by the portability of a DRL navigation stack to different robotic platforms and its capability of online learning. \citet{tai2017virtualtoreal} used Deep Deterministic Policy Gradient (DDPG) to develop a mapless motion planner that enables a robot to navigate  unseen real environments with obstacles. \citet{Jesus2021} replaced the DDPG with the Soft Actor-Critic~\cite{haarnoja2018soft} (SAC) DRL model and showed its efficacy through simulation. %
\citet{long2018optimally} utilized Proximal Policy Optimization (PPO) \cite{schulman2017proximal} to develop a multi-robot collision avoidance policy. The PPO agent directly maps raw sensory data to steering commands. The authors validated the policy in various simulated environments.
\citet{Kulhanek2019} developed a camera-based navigation stack by extending a version of the batched A2C algorithm~\cite{majid2021deep}. The A2C agent was deployed on a real robot for validation \cite{kulhanek2020visual}. 
For Covy, we experimented with DDPG and SAC, a deterministic and probabilistic DRL model, and developed a hybrid navigation stack to overcome some of the observed limitations in a pure DRL approach. 


\section{System Overview}
\subsection{Hardware}
\noindent Figure \ref{fig:robot_overview} shows Covy's hardware and software stack, which are explained next. 

\noindent\textbf{Main CPU.} 
\begin{table}
    \centering
    \caption{The main specifications of the embedded hardware for edge AI taken into account.}
 \begin{threeparttable}
    \begin{tabular}{l| p{2.7cm}| p{2.7cm}| p{2.7cm} | p{2.7cm} }
         \toprule  
          \textbf{Specifications} &  \textbf{Raspberry Pi 4 + \newline Intel Movidius} & \textbf{Raspberry Pi 4 + \newline Coral  Accelerator}  & \textbf{Jetson Nano} & \textbf{Jetson Xavier NX} \\
          \midrule
         GPU  & Up to 150 GFLOPS\tnote{1} & Up to 4 TOPS\tnote{2} (INT8) & 472 GFLOPS (FP32)  &  21 TOPS (INT8)\\
          \midrule
         Accelerator & Broadcom Video Core VI (32-bit) + Myriad X VPU & Broadcom Video Core VI (32-bit) + Google Edge TPU coprocessor  & 128 CUDA cores NVIDIA Maxwell & 384-core NVIDIA Volta GPU with 48 Tensor Cores\\
          \midrule
         CPU & Quad-core ARM CortexA72 64-bit @ 1.5 GHz & Quad-core ARM CortexA72 64-bit @ 1.5 GHz  & Quad-core ARM Cortex-A57 MPCore processor & 6-core NVIDIA Carmel ARMv8.2 64-bit CPU 6MB L2 + 4MB L3\\
          \midrule
         Memory & 8\,GB LPDDR4 + 4\,GB LPDDR3  & 8\,GB LPDDR4  & 4\,GB 64-bit LPDDR4, 1600\,MHz 25.6\,GB/s & 16 GB 128-bit LPDDR4x @ 1866\,MHz 59.7\,GB/s\\
          \midrule
         Storage & Micro-SD & Micro-SD & Micro-SD or 16 GB eMMC 5.1 & Micro-SD or 16 GB eMMC 5.1 \\
          \midrule
         Price & 100 \texteuro & 100 \texteuro & 100 \texteuro & 326 \texteuro \\ 
         \bottomrule 
    \end{tabular}
         \begin{tablenotes}
           \item [1] GFLOPs: a computer system that is capable of performing one billion floating-point operations per second.
           \item [2] TOPS: trillion operations  per second.
         \end{tablenotes}
    \end{threeparttable}
    \label{tab:hw_com}
\end{table}

Covy is based on the Jetson single-board computer (JSBC) family. 
We favored this option because (i) the JSBC family provides multiple processors with increasing CPU and GPU capabilities, and (ii) code portability between these processors is relatively easy. Our experiments were conducted with Jetson Nano and Jetson Xavier NX. Table~\ref{tab:hw_com} compares the used processors to other low-cost options.

\noindent\textbf{Vision Sensor.} 
An RGB-D camera captures RGB images and their depth information on a per-pixel basis \cite{Zhihan2020}. As such it is a natural choice for estimating the 3D coordinates of objects in images. The two most suitable depth cameras for our applications are the ZED2~\cite{zed2} and the Intel RealSense D435i~\cite{agronomy2021depth}. The ZED2 has twice the depth range of the D435i but also costs twice as much. We chose to work with the lower-cost D435i which has a nominal depth range of \SI{10}{\meter}.

\noindent\textbf{Robot body.}
We selected the JetBot AI kit~\cite{jetbotWaveshare} for developing our robot due to its low price ($\approx$ 300 \texteuro) and its growing usage for AI tasks \cite{kawakura2020deep,Raman2018,garcia2021urban}. JetBot is a differential-driven vehicle with an IMX219 8MP camera mounted to the front. It is powered by an NVIDIA processor, allowing it to run AI tasks such as facial recognition, object tracking, auto line following, and collision avoidance. As we target indoor environments we equipped JetBot with a 2D LiDAR~\cite{rplidara2} to enable it to navigate autonomously in buildings.

\subsection{Software}
Covy's software architecture is divided into two main modules: breach detection and navigation. These modules are further sub-divided into several ROS (Robot Operating System) nodes~\cite{ros2009}. 
\subsubsection{Breach Detection}
\begin{figure}  [t]
    \centering
    \includegraphics[width=.9\columnwidth]{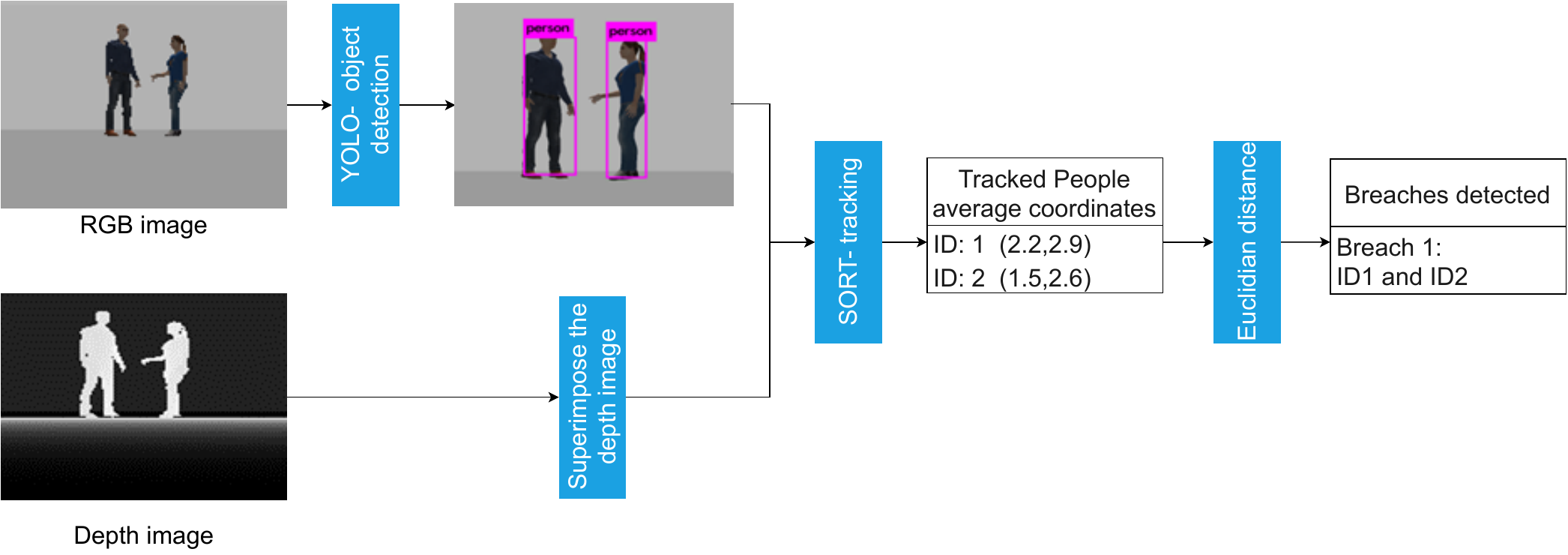}
    \caption{The RGB-D breach detection algorithm.}
    \label{fig:RGBD_overall}
\end{figure}
To localize people in the scene, Covy processes an image twice: first, it processes an RGB-D image to localize nearby people, and second, it analyzes only an RGB image to estimate the locations of faraway people.

\textit{RGB-D based Localization}:- Covy uses YOLOv3~\cite{YOLO2020Tiny} -- a real-time object detection algorithm that features 53 convolutional neural network layers with residual connections -- to detect the 2D coordinates of people in the scene. 
YOLOv3 ROS (ROS implementation of YOLOv3) \cite{redmon2018yolov3} takes an RGB image as input and publishes three ROS topics: bounding boxes around identified objects, their categories, and confidence scores. To determine the 3D coordinates of the identified people, Covy obtains the depth data as a point cloud from the Intel RealSense camera. 

To fuse the depth data with the RGB image, the centers of the boxes surrounding people are calculated. Then the depth points closest to theses centers are chosen to get the 3D coordinates of people in the image. The obtained 3D coordinates are then sent to the SORT (Simple Online and Real-time Tracking) algorithm~\cite{Bewley2016} to track people across multiple images. 
SORT utilizes the position and size of the bounding box for motion estimation and data association using Kalman filter and the Hungarian method, respectively. 
%
Then, it outputs unique IDs for each identified pedestrian. After tracking the identified people for 20 frames,
their 3D coordinates are averaged and the inter-person distance is calculated using the Euclidean distance measure.
If the distance between two individuals is less than a prespecified specified threshold (e.g., \SI{1.5}{\meter}) Covy reports a breach. Figure~\ref{fig:RGBD_overall} visualizes the RGB-D based breach detection process. This process is repeated pairwise for all the detected individuals. The result is a list of breaches containing the different groups of non-compliant pedestrians. Covy determines the largest group, computes its middle coordinates, and sends these coordinates to the navigation module.

\textit{RGB based Localization}:-
Since depth information is only available at a short-range (our experiments show an effective range of \SI{6}{\meter}), when no individuals are detected Covy switches to MonoLoco~\cite{bertoni2019monoloco} for long-range scanning (i.e., up to \SI{20}{\meter}). 
\begin{figure} [t]
    \centering
    \includegraphics[width=.9\columnwidth]{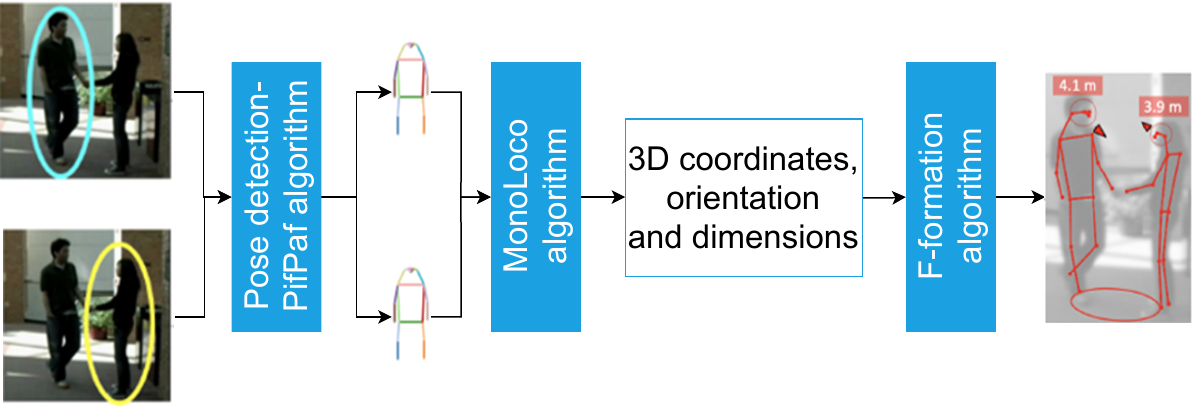}
    \caption{An RGB breach detection algorithm \cite{bertoni2021perceiving}}
    \label{fig:rgb_breach}
\end{figure}
MonoLoco processes an RGB image as a set of 2D joints using two pose detectors: Mask R-CNN~\cite{he2018mask}, which works top-down, and OpenPifPaf~\cite{kreiss2019pifpaf}, which works bottom-up. The MonoLoco algorithm then takes these 2D joints as input and outputs the 3D locations, orientations, and dimensions of the detected people together with localization uncertainty. 
For that, the algorithm uses six fully-connected neural network (DNN) layers of 256 nodes each.  The DNN uses dropout after every fully connected layer and includes batch normalization and residual connections. The output values are analyzed to discover F-formations~\cite{setti2015f}---spatial patterns constructed during interactions between two people or more---and evaluate social distancing breaches (Figure~\ref{fig:rgb_breach}).
Finally, the system publishes an approximation of each pedestrian's \textit{x, y, z} coordinates identified in the image along with their status as breachers or not. The navigation module then guides Covy towards the largest breach density. Using this compound procedure Covy doubles the effective range of Intel RealSense at no additional costs.

\subsubsection{Navigation} 

We have experimented with a deterministic and probabilistic DRL algorithm to develop  Covy's navigation stack. Namely, we implemented the Deep Deterministic Policy Gradient (DDPG)\cite{tai2017virtualtoreal} and Soft Actor-Critic (SAC)  \cite{Xiang2019,Jesus2021} algorithm. The implementation of the said algorithms requires the definition of the state space, action space, network architecture, and reward function which we specify next.

\textit{State Space:-} 
 The environment is observed through 10 laser rays emitted by the LiDAR from -90\degree to 90\degree in front of the robot. 
These measurements are combined with the angular and linear velocity and the relative position and angle of the robot to the target. These values are grouped in vector to  form the input state to a DRL agent. 


\textit{Action Space:-} Both the DDPG and SAC agents have actor-critic network architectures that act in a continuous action space~\cite{lillicrap2019continuous}. The action space has two dimensions: the angular and linear velocities. The angular and linear velocities are limited to [-2,2] rad/s and [0,0.2] m/s for smooth navigation.

\textit{Deep Deterministic Policy Gradient Network (DDPG):-} 
\begin{figure}[t]
    \centering
    \includegraphics[width=.85\columnwidth]{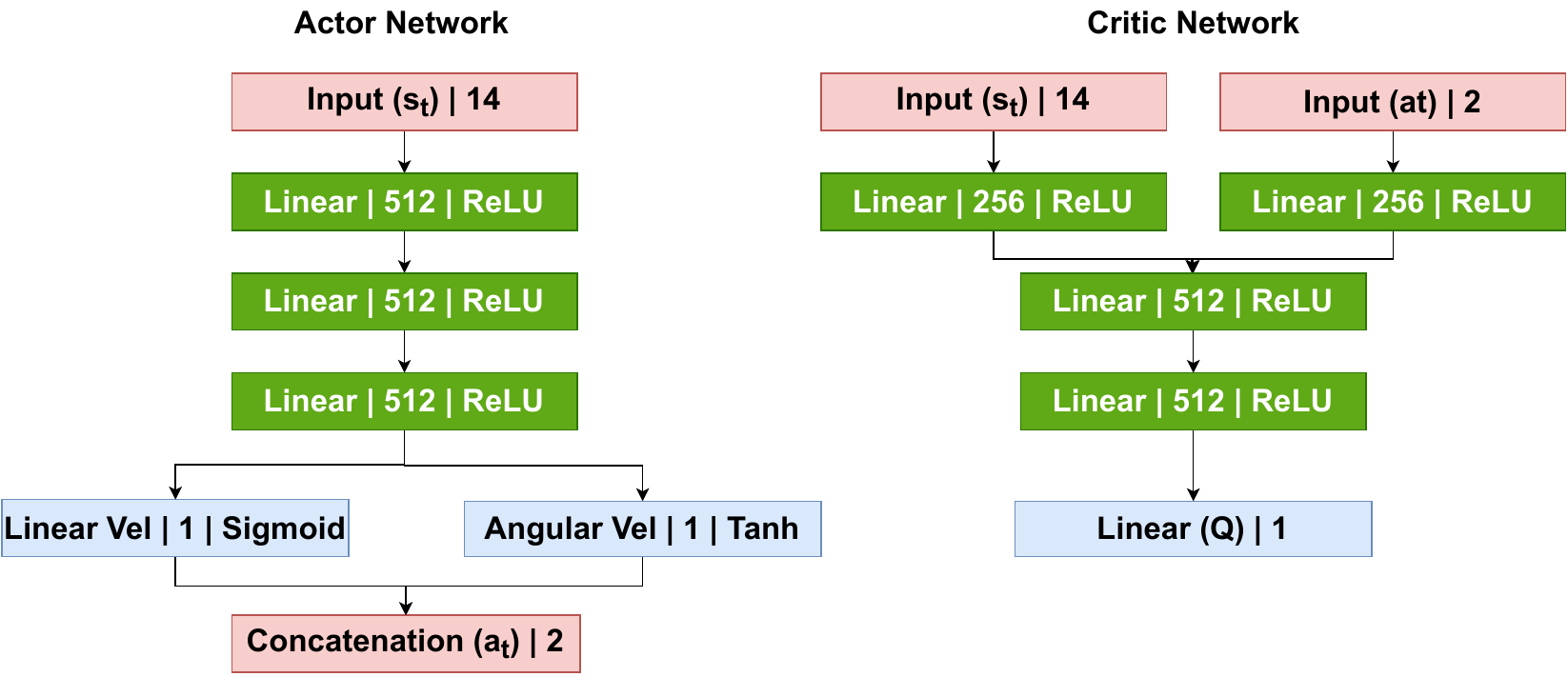}
    \caption{DDPG network structure \cite{tai2017virtualtoreal}.}
    \label{fig:ddpg_network}
\end{figure}
The DDPG \cite{lillicrap2019continuous} is an actor-critic DRL agent. Its actor network consists of three fully-connected neural network layers with 512 nodes each. A rectified linear unit (ReLU) activation follows each layer. The output layer produces two action parameters representing the robot's linear and angular velocities. A  hyperbolic tangent $tanh$ activation function is applied to the angular velocity to limit its range to [-2,2] rad/s, and a $sigmoid$ activation function  keeps the range of the linear velocity within [0,0.2] m/s. 
The critic-network takes as input a pair of a state and action vectors and outputs their associated Q value. Its architecture is similar to that of the actor network except for the first hidden layer which is split into two 256-neurons layers\footnote{Other implementations may use a single 512 neurons layer. Based on our experiments this implementation seemed to give better performance and that is why we chose the architecture with the split first hidden layer.} (Figure \ref{fig:ddpg_network}). Our implementation of the DDPG agent is based on those proposed in \cite{tai2017virtualtoreal,Jesus2021} with minor modifications to the critic network.

\textit{Soft Actor-Critic Network (SAC):-}  
\begin{figure} [t]
    \centering
    \includegraphics[width=.85\columnwidth]{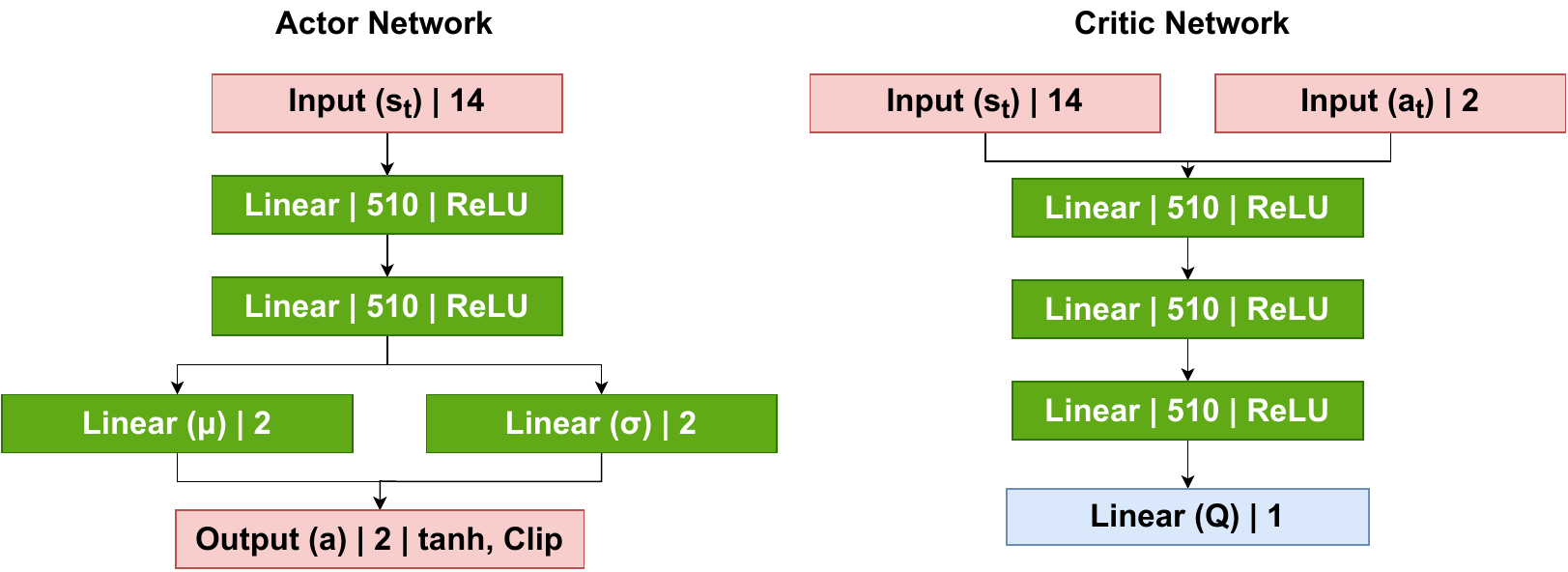}
    \caption{SAC network structure \cite{Xiang2019, haarnoja2019soft}. Each layer includes its type, dimension, and activation function. }
    \label{fig:sac_network}
\end{figure}
SAC consists of three networks: an actor and two critic networks. The critic networks are trained independently, and  the minimum of their Q-values is then used to update the actor network \cite{haarnoja2019soft}.  The actor is composed of 2 fully connected hidden layers with 510 neurons each (Figure \ref{fig:sac_network}). It generates a mean and log standard deviation which are used to output the angular and linear velocities commands. A hyperbolic tangent $tanh$ activation function and the clip operation are applied to limit the linear and angular velocity in the range [0,0.2] m/s and  [-2,2] rad/s, respectively.
A critic network takes as input the current state and action and uses three fully-connected hidden layers to process the associated Q-value. Our implementation of SAC is based on \cite{Xiang2019}; however,  we omitted the value network as recommended in \cite{haarnoja2019soft}.

\textit{Reward function:-} 
A reward signal is what enables a DRL agent to learn. We used 
the following reward function to train our DRL agents (i.e., DDPG and SAC) to navigate to a destination in indoor environments,
$$
r(s_t, a_t) = \begin{cases}
 & r_a \text{ if } D_t < T \\ 
 & r_c \text{ if } L_t < min_c\\ 
 & r_{d1}(D_{t-1}-D_t)\text{ if } (D_{t-1}-D_t) > 0 \\ 
 & r_{d2} \text{ if } (D_{t-1}-D_t) \leq 0 
\end{cases}
$$
Covy receives a large positive reward if it reaches the goal, a large negative reward if it collides\footnote{more precisely, if the LiDAR reading, $L_t$, is smaller than a certain minimum, $min_c$, the robot is regarded to be collided and gets a large negative reward}, a positive reward proportional to its progress towards the goal after each steps, and a constant negative reward otherwise. 
Through such a reward and punishment system, the robot learns to navigate from start to destination while avoiding obstacles. Also, it learns to overcome local minima. For example, if the robot does not receive a large positive reward when reaching the goal, it may learn to crash into the nearest wall to get the smallest negative reward.

\textit{Hybrid Autonomous Navigation:-}
When we deployed the DRL navigation stack on the physical robot, we noticed that the LiDAR odometry---which refers to determining the robot's position relative to its starting point using LiDAR readings---often gets lost.
Whenever this happens, the navigation fails and the robot is unable to reach the target.
To counter this unwanted behavior, we developed a hybrid navigation stack that combines DRL and the Adaptive Monte Carlo Localization (AMCL) algorithm  \cite{Dieter1999}. The AMCL is a probabilistic algorithm that localizes a robot on a given map. To recover from any failure mode that may happen due to LiDAR odometry issues, Covy obtains the robot pose estimation from AMCL and compares it to the DRL estimation. If there is a significant difference, Covy reinitializes its pose using AMCL coordinates. Comparing the two estimations frequently slows down the computation; therefore, this comparison happens every X steps (in our implementation every 20 steps). 




\begin{figure} [t]
    \begin{subfigure}[b]{.245\columnwidth}
        \centering
        \includegraphics[width=\columnwidth]{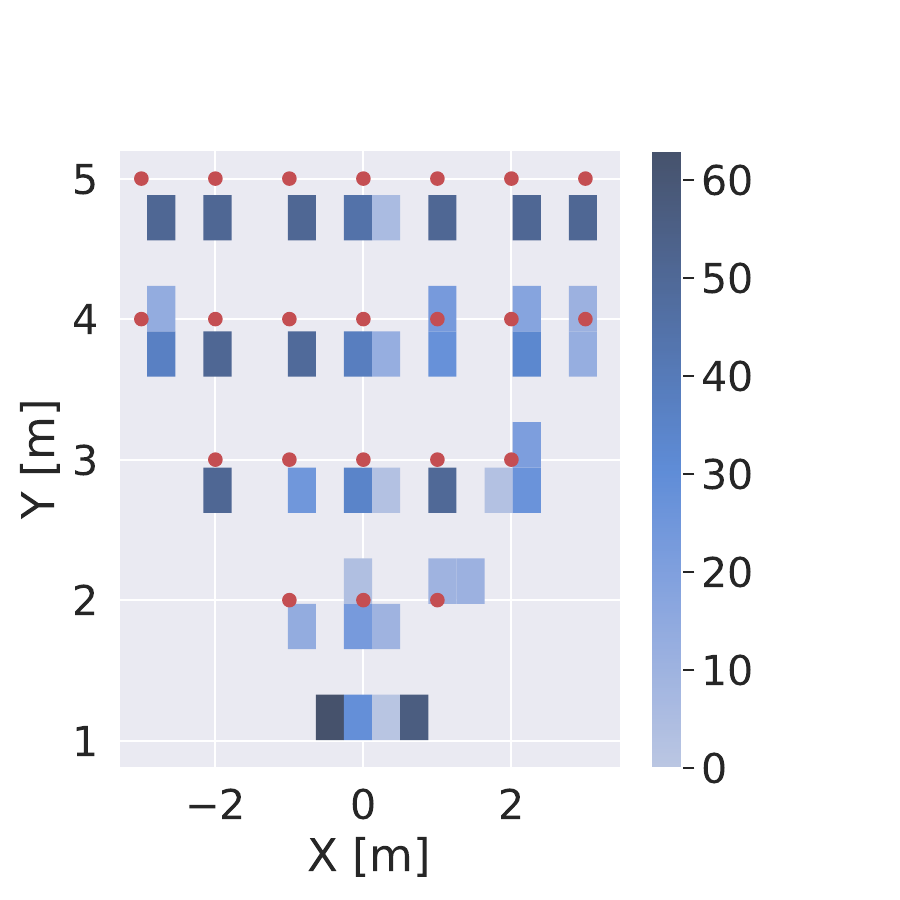}
        \caption{RGB-D simulation}
        \label{fig:heatmapRGBDsim}
    \end{subfigure} \hfill
    \begin{subfigure}[b]{.245\columnwidth}
        \centering
        \includegraphics[width=\columnwidth]{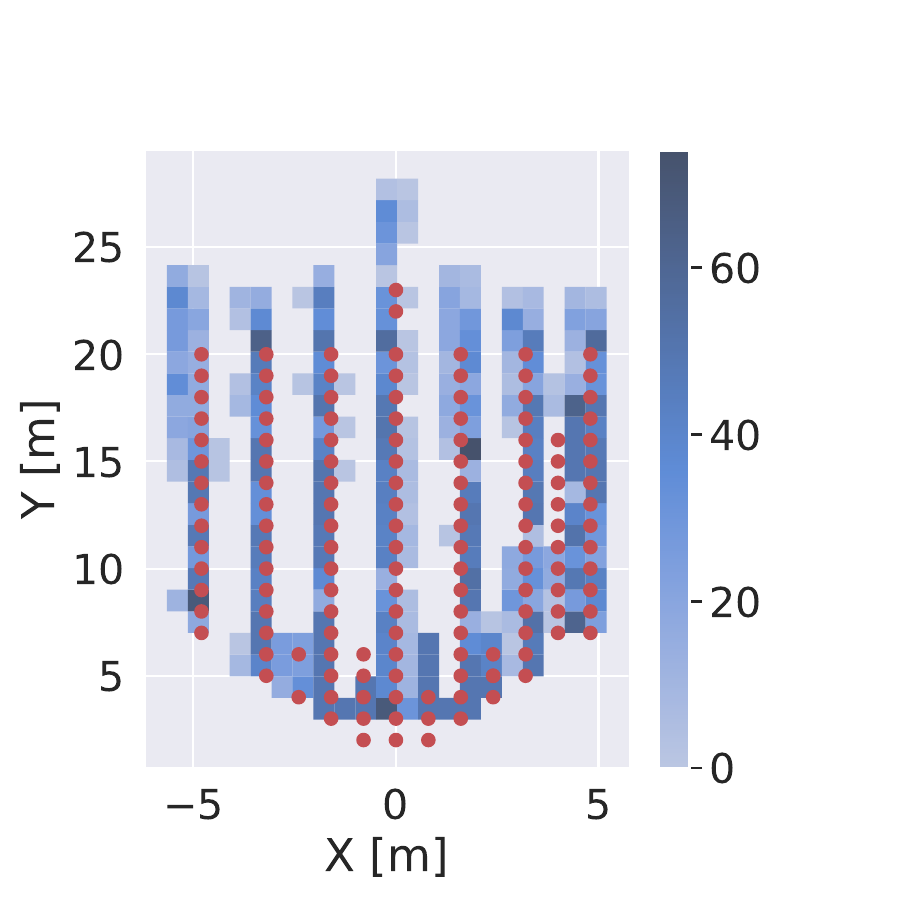}
        \caption{RGB simulation}
        \label{fig:heatmapRGBsim}
    \end{subfigure} \hfill
    \begin{subfigure}[b]{.245\columnwidth}
        \centering
        \includegraphics[width=\columnwidth]{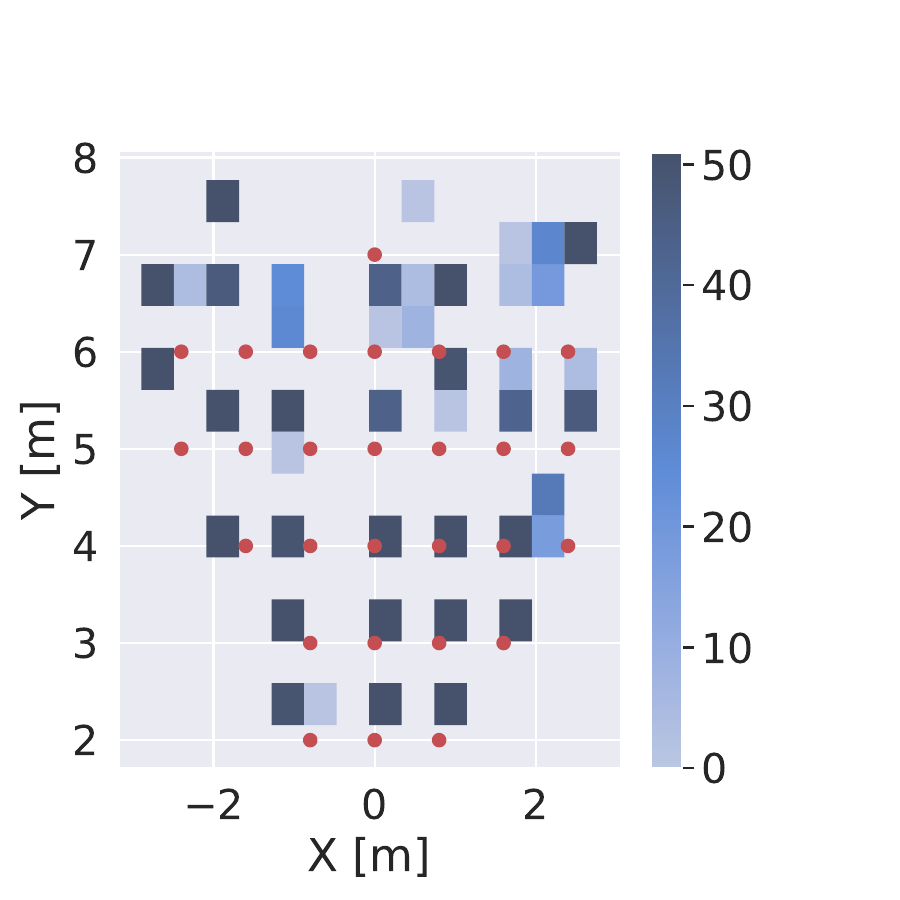}
        \caption{RGB-D Real}
        \label{fig:heatmapRGBDreal}
    \end{subfigure} \hfill
    \begin{subfigure}[b]{.245\columnwidth}
        \centering
        \includegraphics[width=\columnwidth]{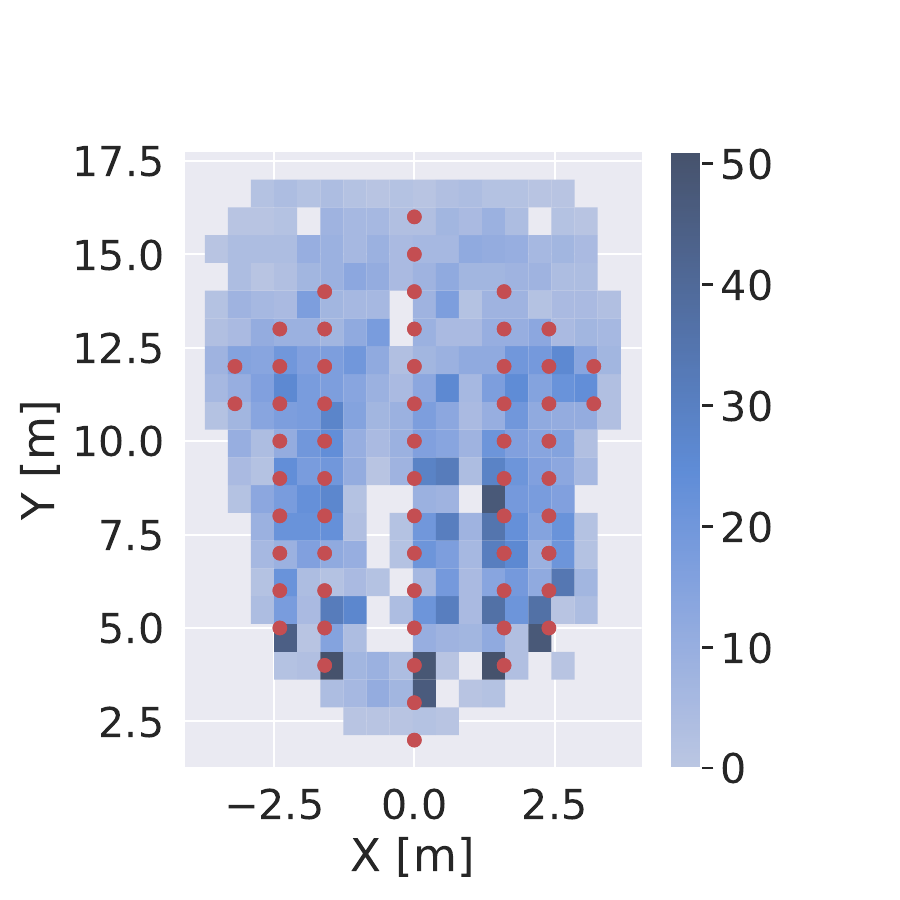}
        \caption{RGB Real }
        \label{fig:heatmapRGBreal}
        \end{subfigure} \hfill
     \caption{Heatmap showcasing the ground truth measurements (red circles) versus the approximated ones (blue squares).}
    \label{fig:heatmap}
\end{figure}

\section{Experimental Evaluation}
Covy's vision and navigation systems are evaluated next.  
\subsection{Vision System}

Two sets of experiments were conducted to evaluate Covy's vision system: person's location estimation accuracy and classification accuracy of breaches in social distancing. Each experiment was done in a simulated and real environment. 
For the person's location estimation, we moved the person away from the camera by 1\,m for every new experiment and took 50 pictures at each location. Additionally, we sampled a few other coordinates by moving the pedestrian vertically by 0.8\,m. We repeated this experiment until the camera could not detect any pedestrian, which resulted in ranges of 23 and  16\,m for the RGB camera and 5 and 7\,m for the RGB-D camera in the simulated and real environment, respectively. 
The red circles in Figure \ref{fig:heatmap} show the exact coordinates at which a person stood.  For the breach detection accuracy, we placed two people at various coordinates in the scene while ensuring they were in the camera's field of view and took 50 pictures at each location. To get balanced data, we ensured that they breached the social distancing rule in half of the experiment trials. 
For detecting people using the RGB-D camera, we used the YOLOv3 model~\cite{redmon2018yolov3} trained on the COCO dataset \cite{lin2015microsoft}. For the RGB-based detection, we used the MonoLoco model~\cite{bertoni2021perceiving}, trained on the KITTI dataset \cite{Geiger2013IJRR}.
Finally, we profiled the execution of these algorithms the Jetson Nano and Jetson Xavier NX processor.

\subsubsection{ Localization} 

\begin{figure} [t]
    \begin{subfigure}[b]{.24\columnwidth}
        \centering
        \includegraphics[width=\columnwidth]{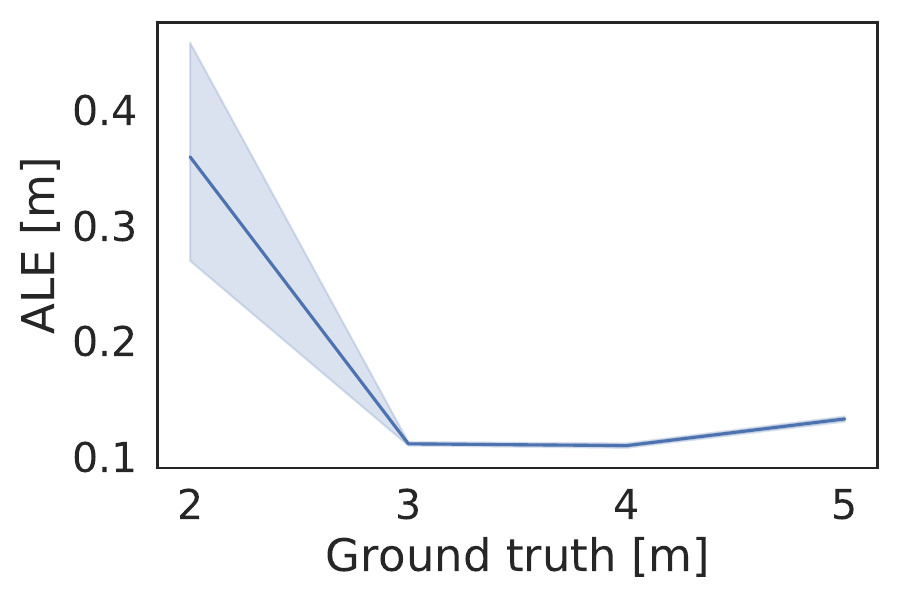}
        \caption{RGBD simulation}
        \label{fig:ale_rgbdsim}
    \end{subfigure} \hfill
    \begin{subfigure}[b]{.24\columnwidth}
        \centering
        \includegraphics[width=\columnwidth]{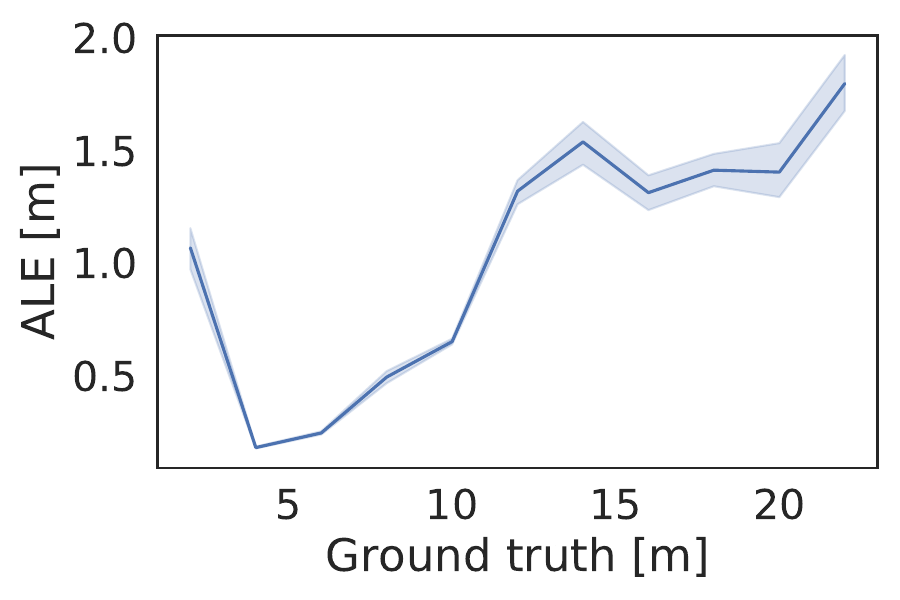}
        \caption{RGB simulation}
        \label{fig:ale_rgbsim}
    \end{subfigure} \hfill
    \begin{subfigure}[b]{.24\columnwidth}
        \centering
        \includegraphics[width=\columnwidth]{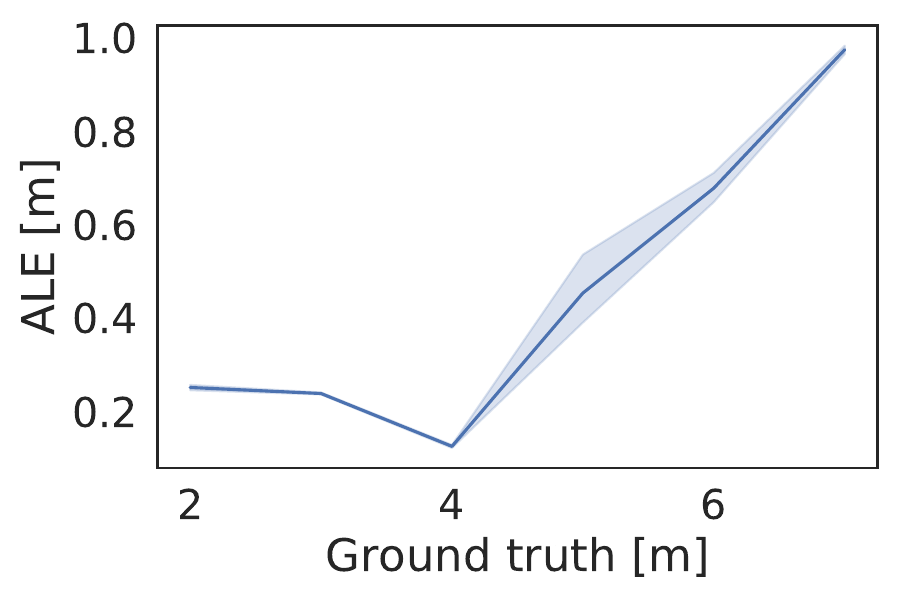}
        \caption{RGBD Real}
        \label{fig:ale_rgbdreal}
    \end{subfigure} \hfill
    \begin{subfigure}[b]{.24\columnwidth}
        \centering
        \includegraphics[width=\columnwidth]{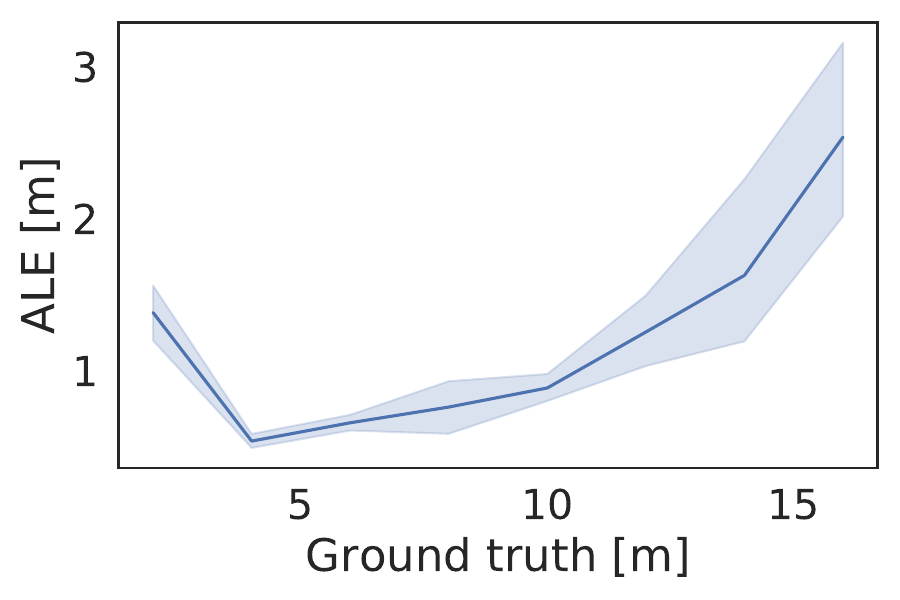}
        \caption{RGB Real }
        \label{fig:ale_rgbreal}
        \end{subfigure} \hfill
     \caption{Average localization error (ALE) with confidence interval as a function of distance --  RGBD and RGB based system.}
    \label{fig:localization_error}
\end{figure}

Figure~\ref{fig:localization_error} shows the average localization error (ALE) of Covy's vision RGB-D and RGB localization methods. Two main conclusions we drew from these results are as follows. 

First, from Figure~\ref{fig:ale_rgbdsim} and \ref{fig:ale_rgbdreal}, we observe that there is a noticeable difference in the ALE patterns of the simulated and physical RGB-D camera. 
We conjecture that this discrepancy is due to the difficulty of simulating how the Intel RealSense camera~\cite{realsense} estimates depth. It uses active illumination for measuring distances to objects. Precise simulation of active illumination requires accurate ambient light simulation which is hard. On the other hand, the results of the simulated and real RGB camera are more consistent (Figures~\ref{fig:ale_rgbsim} and ~\ref{fig:ale_rgbreal} ). This is expect as the input to both is a pure RGB image.

Second, when assessing the ALE in Figure~\ref{fig:ale_rgbdreal}, we notice that the RGB-D camera can detect pedestrians standing at less than $5\,m$ with an ALE $< 0.4\,m$. The RGB system, however, reaches its minimum ALE of $\approx 0.4\,m$ at $5\,m$ and then the error goes up again to an average of $\approx 2.5\,m$ at $16\,m$ (\ref{fig:ale_rgbreal}). Furthermore, the variation in the coordinates estimation of the RGB method is higher than that of the RGB-D one. Consequently, Covy is less certain about the locations of faraway people; but, the accuracy increases as the distance between Covy and people shrinks. 
%
In conclusion, at short distances, i.e. $\leq 5\,m$, the RGB-D detection algorithm is the preferred choice, while from $5\,m$ onwards, the RGB detection method is a better fit. Covy capitalizes on both to detect breaches up to $16\,m$ away and approach the offenders safely.

%
\subsubsection{Breaches Classification Accuracy} 

\begin{figure} [t]
    \begin{subfigure}{.24\textwidth}
        \centering
        \includegraphics[width=\textwidth]{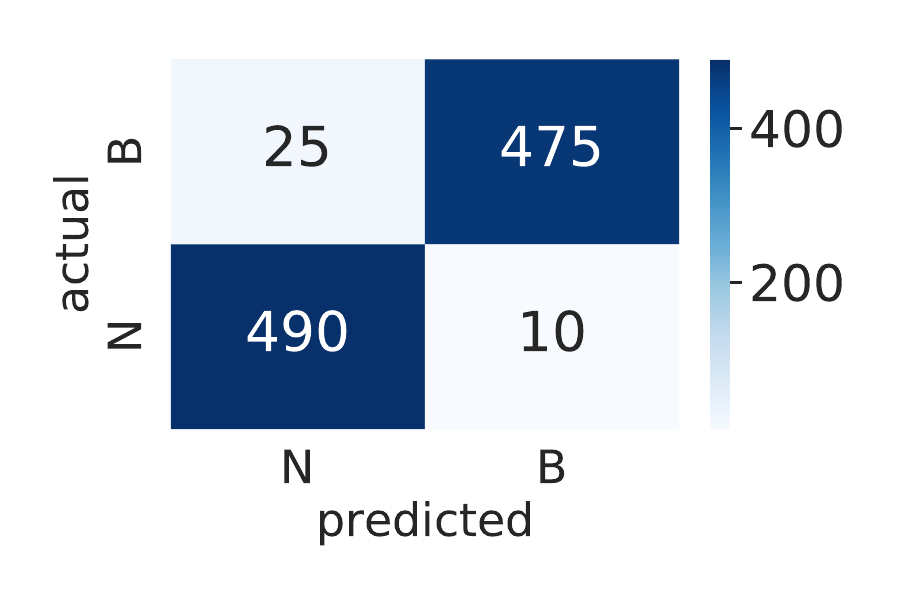}
        \caption{RGB-D  simulation}
      \label{fig:rgbdsim}
    \end{subfigure} 
    \begin{subfigure}{.24\textwidth}
        \centering
        \includegraphics[width=\textwidth]{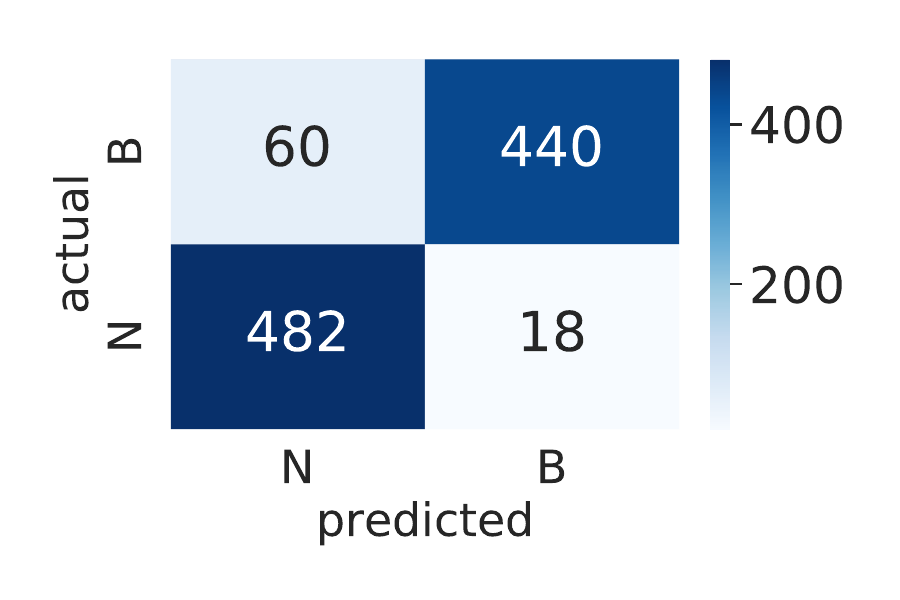}
        \caption{RGB  simulation}
        \label{fig:rgbsim}
    \end{subfigure} 
    \begin{subfigure}{.24\textwidth}
        \centering
           \includegraphics[width=\textwidth]{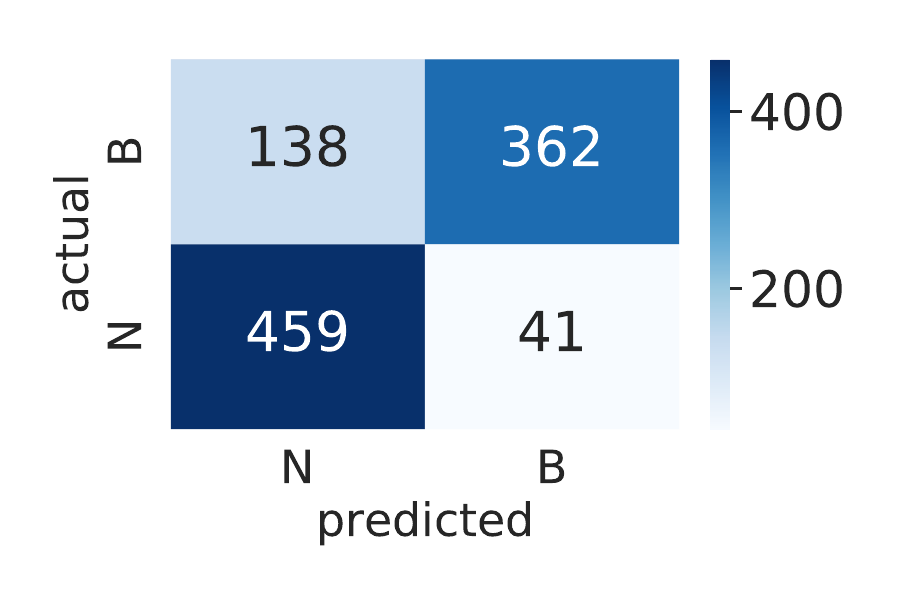}
        \caption{RGB-D  real env. }
      \label{fig:rgbdreal}
    \end{subfigure} 
    \begin{subfigure}{.24\textwidth}
        \centering
        \includegraphics[width=\textwidth]{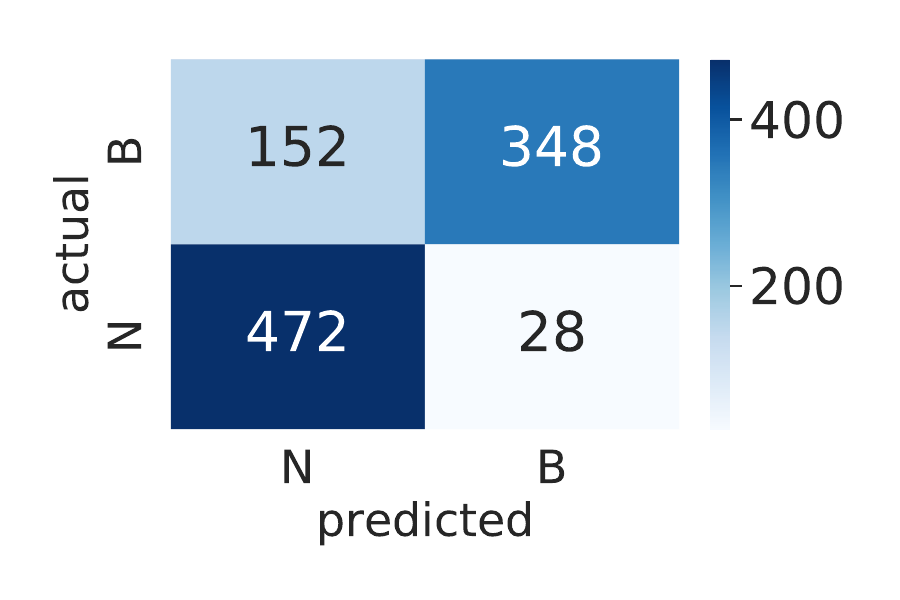}
        \caption{RGB  real env.}
        \label{fig:rgbreal}
    \end{subfigure} 
     \caption{Accuracy in Detection of Breaches Using Covy's Vision Modules (RGB and RGB-D) in simulated and real environment. N: no breaches and B: breaches.}
    \label{fig:overall}
\end{figure}

This experiment compares the performance of the RGB-D and RGB-based vision systems in detecting social distancing violations. We treat this problem as a binary classification task and evaluate the detection accuracy, recall, and precision. Figure \ref{fig:overall} shows the confusion matrices of the classification, and Table \ref{tab:classification} compares the said metrics between the two methods. 

\textit{Simulation:-}
From Table \ref{tab:classification}, we notice that the accuracy is equal to 96\% and 92\% for the RGB-D and RGB systems with high precision and recall. 
Both methods can therefore accurately classify social distancing breaches in simulation. However, the recall of the RGB method is lower than that of the RGB-D one by $7\%$. In other words, the RGB vision system classifies more breaches as safe compared to the RGB-D counterpart. The reason is that, the RGB system has about three times the detection range of the RGB-D camera but the variance in the localization error increases significantly for large distances ($> 13\,m$, Figure \ref{fig:ale_rgbsim}). 
Hence, when running the experiments, we noticed that a large number of false negatives were detected for distances above $15\,m$.

\begin{figure} [t]
    \begin{subfigure}[b]{.44\textwidth}
        \centering
        \includegraphics[width=.9\columnwidth]{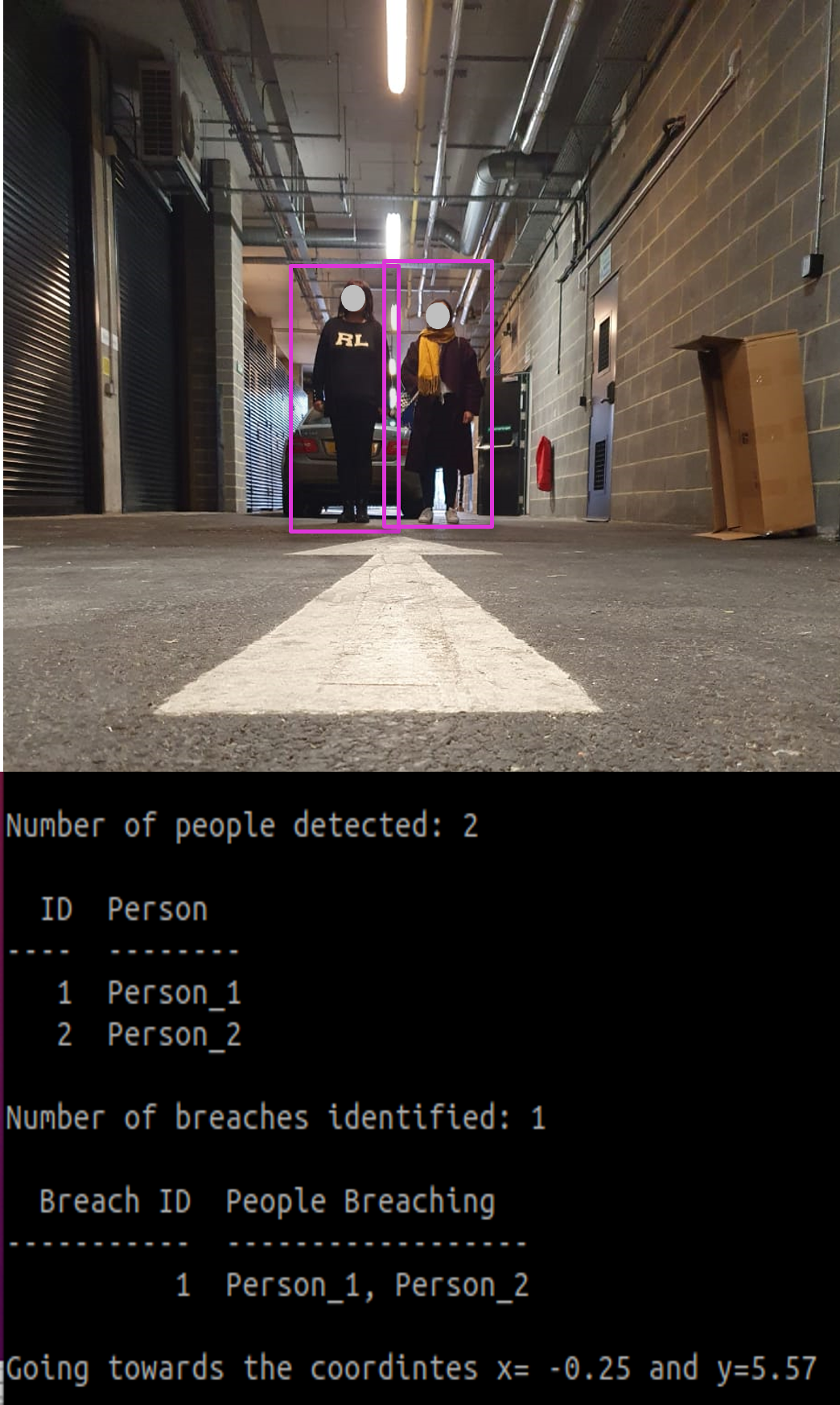}
        \caption{RGBD qualitative results}
      \label{fig:rgbdqual}
    \end{subfigure} 
    \begin{subfigure}[b]{.44\textwidth}
        \centering
        \includegraphics[width=.9\columnwidth]{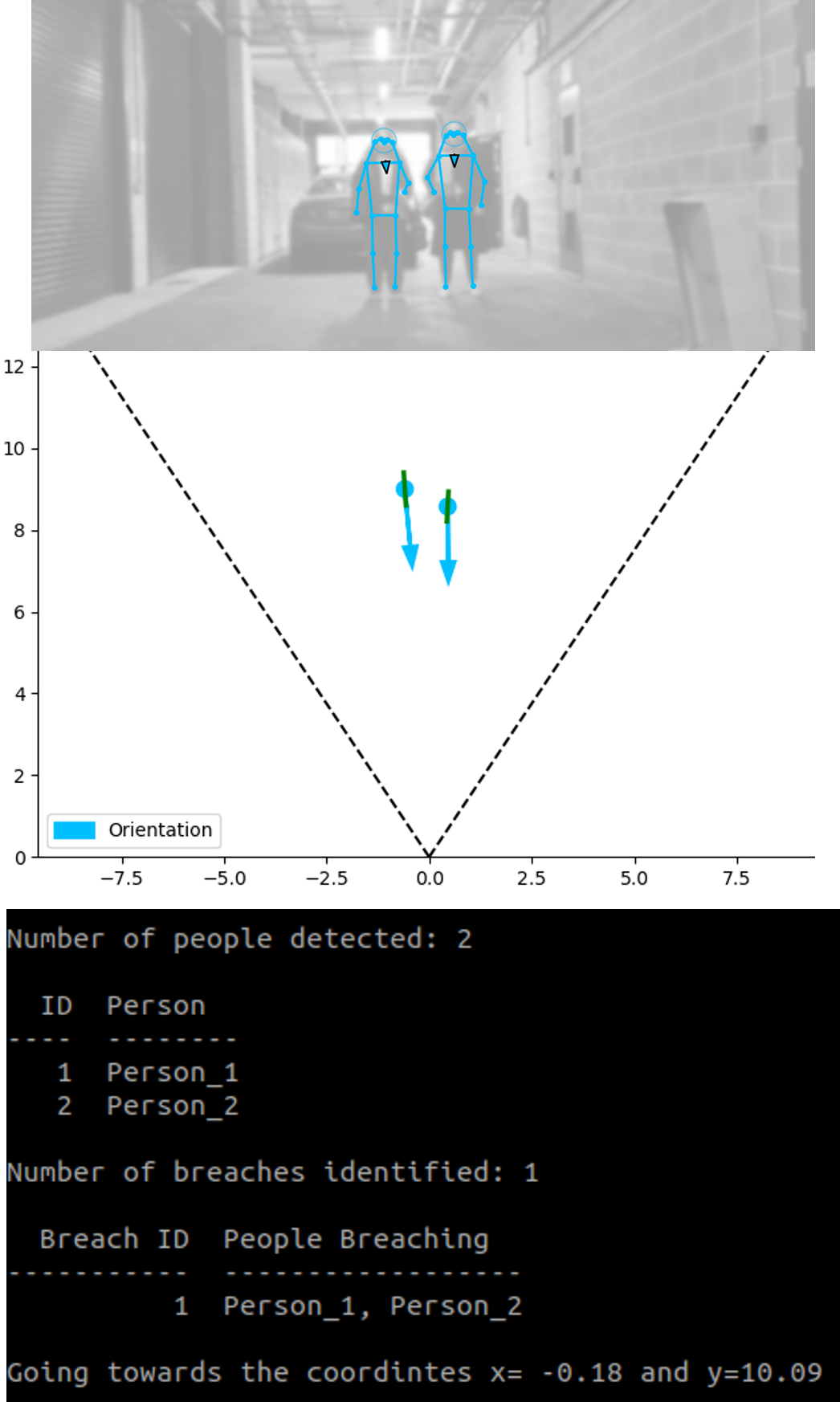}
        \caption{RGB qualitative results}
        \label{fig:rgbqual}
    \end{subfigure} \hfill
     \caption{Example of Covy's vision modules (i.e., RGB and RGB-D) and detection accuracy of breaches.}
    \label{fig:qualitative}
\end{figure}

\textit{Reality:-}
Similarly, the classification precision is high in the real environment $(\geq 90\%)$ (Table \ref{tab:classification}). However, the accuracy and recall fall off. This decrease is due to the negative effect of the real environment and its surrounding on the performance of our systems. 
Additionally, we noticed, during the experiments, that many misclassifications happen at the limits of the cameras' field of view (i.e., where the ALE is high). 
Overall, both algorithms maintained accuracy of 82\% in realistic conditions. Qualitative results of both methods are shown in Figure \ref{fig:qualitative}. Combined, these results show the promising ability of Covy in recognizing breaches in social distancing.

\begin{table}
    \begin{tabular}{l|c|c|c|l}
        \toprule
        \textbf{Method}  & \makecell{ \textbf{Accuracy}  (\%)}  & \makecell{ \textbf{Precision} (\%)}     & \makecell{\textbf{Recall} (\%)} & \textbf{Environment}\\ 
        \midrule
        RGB-D    & 96      & 98          & 95  & Simulation \\
        RGB    & 92      & 96          & 88 & Simulation  \\
        RGB-D      & 82      & 90          & 72 &  Reality \\
        RGB       & 82      & 93          & 70 & Reality \\ 
        \bottomrule
    \end{tabular}
    \caption{Accuracy of classifying social distancing breaches}
    \label{tab:classification}
\end{table}

\subsubsection{Hardware Performance}
\begin{table*}[t]
\centering
    \begin{tabular}{l|l|c|c|c|c|c|c}
         \toprule  
          \textbf{Hardware} & \textbf{Status} &\textbf{CPU1(\%)} & \textbf{CPU2(\%)} & \textbf{CPU3(\%)} & \textbf{CPU4(\%)}  & \textbf{Memory (GB)} & \textbf{GPU(\%)}\\
          \midrule
                            & Idle            & 18        & 15         & 15       & 14       & 1.6/4.1     & 3\\ 
                    Nano    & RGB detection   & NA        & NA         & NA       & NA       & NA          & NA \\ 
                            & RGB-D detection & 99        & 98         & 99       & 99       & 2.7/4.1     & 83 \\
         \midrule
                      & Idle & 15 & 15 & 13 & 13 & 2.5/8 & 1\\ 
            Xavier NX & RGB detection & 44 & 42 & 36 & 34 & 7.2/8 & 59 \\
            & RGB-D detection & 60 & 50 & 56 & 65 & 3.3/8 & 15 \\
            & Covy's vision & 76 & 63 & 70 & 75 & 7.4/8 & 63 \\
         \bottomrule
    \end{tabular}
     \caption{Performance comparison of breach detection algorithms on different hardware.}
    \label{tab:boards}
\end{table*}

To analyze the computational resources utilization of the RGB and RGB-D methods, we profiled their execution on  Jetson Nano and Jetson Xavier NX.  We measured the CPU, GPU, and memory usage as shown in Table \ref{tab:boards}. 
For each iteration, we measured resource utilization at idle conditions, run the model for a few minutes to let it reach its steady-state conditions, and then took 100 measurements. 

Jetson Nano was able to run the RGB-D detection system based on Yolov3 \cite{redmon2018yolov3}, but reached maximum CPU utilization ($\approx 99 \%$) and very high GPU utilization ($83 \%$). Moreover, Jetson Nano was unable to run the resource-intensive RGB-based detection algorithm which runs multiple deep learning models to estimate the 3D coordinates of people in the scene. 

Jetson Xavier NX, on the other hand, was able to run both systems with acceptable resource utilization. The RGB-based detection method noted the highest GPU utilization (59\%), while the RGB-D system showed the highest CPU usage (65\%). However, the memory consumption of the RGB system was high, averaging 7.2 GB of RAM out of the total 8.

Considering that the individual performances were favorable, we tested Covy's vision algorithm as a whole on Jetson Xavier NX. The overall performance seemed promising, with average utilization of 70\%, 63\%, and 92\% of the CPU, GPU, and RAM, respectively. 

\subsection{Navigation}

We evaluated the performance of the navigation stacks using success rate, failure rate due to collides or deadlocks, and average speed. 

\subsubsection{Experiment setup} 
\begin{figure}
    \centering
    \subfloat[\centering Scenario 1: No obstacles ]{{\includegraphics[width=.32\textwidth]{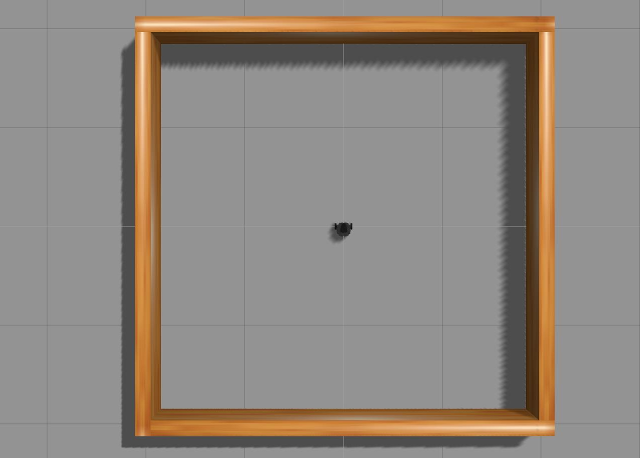} }
    \label{fig:t1}}%
    \hfill
    \subfloat[\centering Scenario 2: Static obstacles ]{{\includegraphics[width=.32\textwidth]{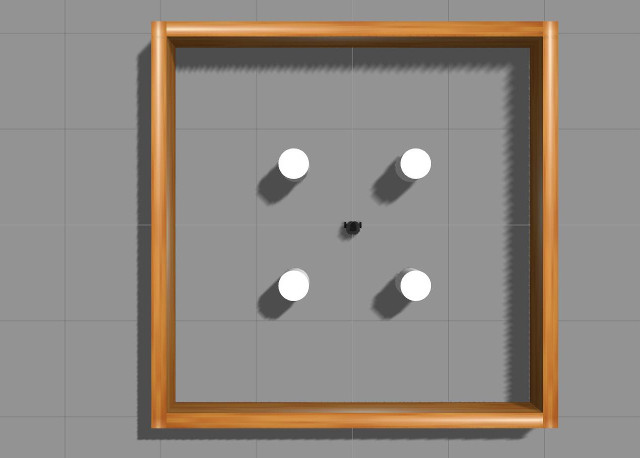} }
    \label{fig:t2}}%
    \hfill
    \subfloat[\centering Scenario 3: Static and dynamic obstacles ]{{\includegraphics[width=.32\textwidth]{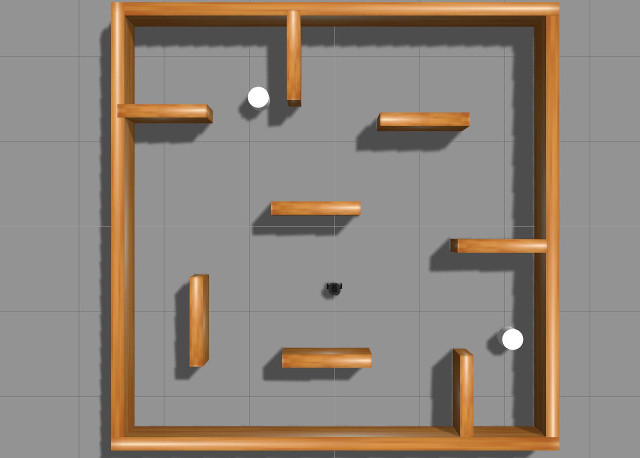} }
    \label{fig:t3}}%
    \caption{Training environments.}
    \label{fig:env_img}
\end{figure}

We trained the DRL models on a computer equipped with an NVIDIA GeForce GTX 1060, 16 GB of RAM, and an Intel Core i7-8750H processor.
We used the three virtual environments shown in Figure \ref{fig:env_img}, to train and test our DRL algorithms using the Jetbot robot model.
The environments consist of a $4\times\SI{4}{\meter}^2$ room-like environment with no obstacles, static obstacles, and dynamic obstacles. The DRL models are trained for 4000 episodes by navigating a Jetbot robot model in these environments. We tested the models capabilities in simulation and reality by sampling 35 random configurations for each environment type. 

\subsubsection{Training} 
\begin{figure} [t]
    \centering
    \includegraphics[width=.8\columnwidth]{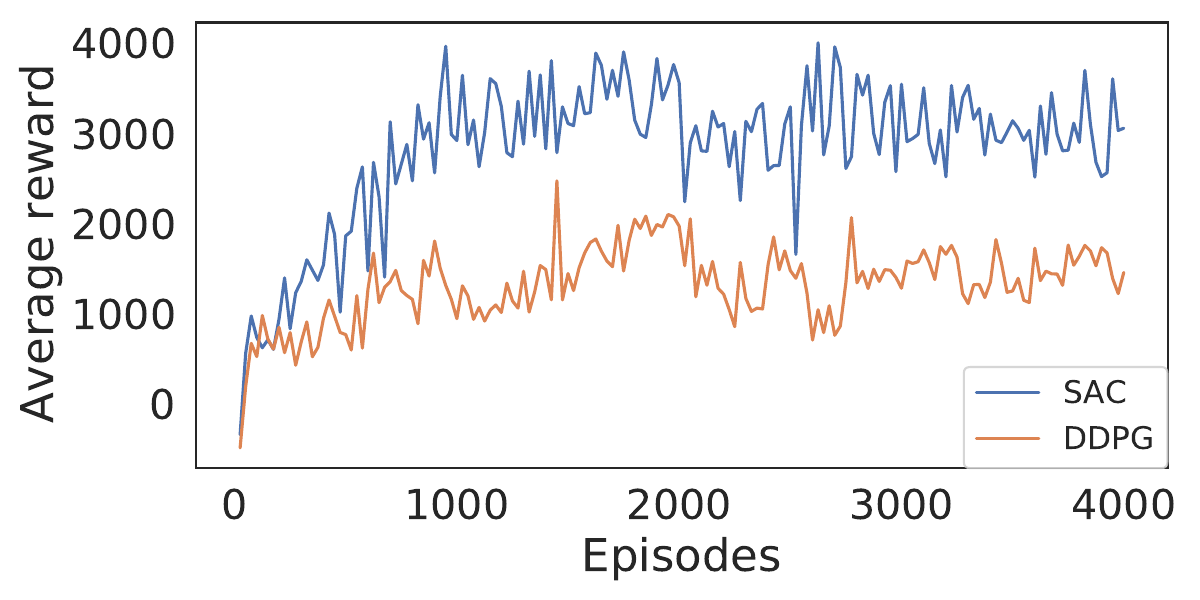}
    \caption{Training in environment with varying obstacles.}
    \label{fig:DDPG_SAC}
\end{figure}

Figure \ref{fig:DDPG_SAC} shows the cumulative rewards obtained by DDPG and SAC during training in an environment with static and dynamic obstacles where each data point represents the average reward over 25 episodes. After $\approx$ 350 episodes, SAC surpasses DDPG in performance constantly. SAC reached a maximum average reward of approximately 3000, double that of the DDPG agent. We hypothesize that as SAC is a stochastic policy, it is able to explore the environment better and therefore collect higher rewards.

\subsubsection{Navigation Accuracy}
 \begin{table*}[t]
\centering
    \caption{Comparison of the thee navigation stacks in simulated and real environments.}
    \begin{tabular}{c |l | c | c | c}
         \toprule  
          \textbf{Env.} & \textbf{Algorithms} & \textbf{Success (\%) } & \textbf{Failures  (collisions $/$ deadlocks) (\%) } & \textbf{Average Speed [m/s]}\\
          \midrule
          \multirow{3}{2em}{ \begin{turn}{90}Simulation \end{turn} }& DDPG & 87 & 13 (8 / 5)  & 0.125\\[5pt]
            & SAC & 92 & 8 (6 / 2) & 0.17\\[5pt]
            & Hybrid navigation & 93 & 7 (7 / 0) & 0.046\\
          \midrule
          \multirow{3}{2em}{ \begin{turn}{90}Reality \ \ \ \  \end{turn} } & DDPG & 74 & 26 (11 / 15) & 0.099\\[5pt]
          & SAC & 80 & 20 (8 / 12)  & 0.13\\[5pt]
          & Hybrid navigation & 90 & 10 (10 / 0) & 0.057\\
         \bottomrule
    \end{tabular}
    \label{tab:nav}
\end{table*}
Table \ref{tab:nav} shows that the thee navigation stacks are performant in simulation with a minimum success rate of  $\geq 87\%$. However, in reality the performances drop by more than $10\%$ for pure DRL approaches and by $3\%$ for the hybrid navigation stack. On the other hand, the hybrid stack is about 2 to 3 times slower than its DRL counterparts. This is expected as the robot needs to keep track of its position on the given map. 
Furthermore, we see that the SAC model performs better than the DDPG model in the real and simulated environment. Additionally, SAC is faster than DDPG in reaching the target: we suspect that the stochastic nature of SAC provides better exploration and allows the model to find more  optimal paths than DDPG, which is a deterministic DRL agent. Contrary to the hybrid approach, both DDPG and SAC can get lost (or end up in a deadlock) in both environments. We conjecture that this is due to the use of LiDAR odometry which can produce inconsistent measurement when the robot moves fast. Overall, our hybrid approach performs the best the cost of slower navigation.

\section{Conclusion and Future Work}
\texttt{Covy} is a robot designed to test a compound vision system and different navigation stacks. The target application is to promote social distancing practice during a pandemic (e.g., COVID-19).
The vision algorithm that Covy uses extends the range of depth of the Intel RealSense D435i camera from an effective range of $\SI{6}{\meter}$ to $16\,m$. 
Covy navigates its surroundings autonomously using a hybrid navigation stack that combines a  DRL agent with the probabilistic localization method AMCL. 
Results from navigating in realistic and simulated environments show that  Covy's hybrid navigation stack is superior to a pure DRL-based one.

To further reduce the costs, next, we are planning to drop the LiDAR and develop a vision-based navigation stack. Then, a swarm of Covy robots shall be developed to target applications such as alerting littering individuals or finding objects of interest in airports or alike environments.

\bibliographystyle{ACM-Reference-Format}
\bibliography{main}

\end{document}